\documentclass[lettersize,journal]{IEEEtran}
\usepackage{amsmath,amsfonts}
\usepackage{algorithmic}
\usepackage{algorithm}
\usepackage{array}
\usepackage[caption=false,font=normalsize,labelfont=sf,textfont=sf]{subfig}
\usepackage{textcomp}
\usepackage{stfloats}
\usepackage{url}
\usepackage{verbatim}
\usepackage{graphicx}
\usepackage{cite}
\usepackage{multirow}
\usepackage[capitalize]{cleveref}
\def\eg{\emph{e.g.}} 
\def\ie{\emph{i.e.}}

\def\etal{\emph{et al.}}
\usepackage{color}
\definecolor{blue}{rgb}{0.0,0.0,1.0}

\begin{document}

\title{FullLoRA: Efficiently Boosting the Robustness of Pretrained Vision Transformers}

\author{Zheng Yuan,~\IEEEmembership{Student Member,~IEEE,}
        % <-this % stops a space
        Jie Zhang,~\IEEEmembership{Member,~IEEE,}
        Shiguang Shan,~\IEEEmembership{Fellow,~IEEE,}
        Xilin Chen,~\IEEEmembership{Fellow,~IEEE}
\IEEEcompsocitemizethanks{
% \IEEEcompsocthanksitem J. Zhang is the corresponding author. \protect\\
\IEEEcompsocthanksitem Z. Yuan, J. Zhang, S. Shan and X. Chen are with the Key Laboratory of AI Safety of CAS, Institute of Computing Technology, Chinese Academy of Sciences (CAS), Beijing 100190, China, and the University of Chinese Academy of Sciences, Beijing 100049, China. E-mail: zheng.yuan@vipl.ict.ac.cn, \{zhangjie, sgshan, xlchen\}@ict.ac.cn.}
% \\ is fragile and will error, could use \hfil\break instead.}% <-this % stops a space
\thanks{Manuscript received April 19, 2005; revised August 26, 2015.}
}


\markboth{Journal of \LaTeX\ Class Files,~Vol.~14, No.~8, August~2021}%
{Shell \MakeLowercase{\textit{et al.}}: A Sample Article Using IEEEtran.cls for IEEE Journals}


\maketitle

\begin{abstract}
  In recent years, the Vision Transformer (ViT) model has gradually become mainstream in various computer vision tasks, and the robustness of the model has received increasing attention. However, existing large models  tend to prioritize performance during training, potentially neglecting the robustness, which may lead to serious security concerns. In this paper, we establish a new challenge: exploring how to use a small number of additional parameters for adversarial finetuning to quickly and effectively enhance the adversarial robustness of a standardly trained model. To address this challenge, we develop novel LNLoRA module, incorporating a learnable layer normalization before the conventional LoRA module, which helps mitigate magnitude differences in parameters between the adversarial and standard training paradigms.
  Furthermore, we propose the FullLoRA framework by integrating the learnable LNLoRA modules into all key components of ViT-based models while keeping the pretrained model frozen, which can significantly improve the model robustness via adversarial finetuning in a parameter-efficient manner.
  Extensive experiments on several datasets demonstrate the superiority of our proposed FullLoRA framework. It achieves comparable robustness with full finetuning while only requiring about 5\% of the learnable parameters. This also effectively addresses concerns regarding extra model storage space and enormous training time caused by adversarial finetuning.
\end{abstract}


\section{Introduction}
\label{sec:intro}
\IEEEPARstart{I}{n} recent years, with the rapid development in the field of computer vision, deep learning models have made profound progress in various tasks, such as image classification~\cite{simonyan2014very, he2016deep, szegedy2016rethinking}, face recognition~\cite{liu2017sphereface, wang2018cosface, deng2019arcface}, and semantic segmentation~\cite{chen2015semantic, chen2018deeplab, chen2017rethinking}. Along with the increasing accessibility of computational resources and the rapid growth in the number of model parameters and training data, Vision Transformer (ViT)-based large models~\cite{dosovitskiy2021an, touvron2021training, liu2021swin} have gradually become the mainstream models for various tasks~\cite{carion2020end, zheng2021rethinking, arnab2021vivit}.

Despite the significant success of large models, there is often more focus on performance during training, which can sometimes lead to neglecting the robustness of the model. Unfortunately, existing work has found that deep learning models are vulnerable to attacks by adversarial examples~\cite{szegedy2014intriguing, goodfellow2015explaining, kurakin2017adversarial, dong2018boosting}. This means that adding minimal, imperceptible perturbations to the input can easily lead to incorrect outputs from the model, whether it is CNN-based~\cite{goodfellow2015explaining, dong2018boosting} or ViT-based~\cite{lovisotto2022give, fu2022patch} models.

Given this situation, we propose a new task setting: \emph{for models that have already been standardly pretrained, we aim to employ a lightweight approach to finetune them, thereby quickly enhancing their robustness against the adversarial examples}. The lightweight aspect we refer to primarily concerns the number of additional model parameters that need to be trained during finetuning, as well as the additional training time required.

Many existing research works have proposed various adversarial defense methods to enhance model robustness, including adversarial training~\cite{madry2018towards, zhang2019theoretically, zhu2023improving, liu2023twins, wang2024pre}, adversarial purification~\cite{nie2022diffusion}, adversarial example detection~\cite{li2020connecting}, and certifiable defenses~\cite{croce2020provable}, among which adversarial training is the most commonly used and effective method. Adversarial training typically involves incorporating adversarial examples into the training set, enabling the model to resist such examples during training and significantly enhancing its robustness. 
Traditional approaches to adversarial training typically involve either training from scratch or fully finetuning from a pretrained model. This process requires updating the entire set of model parameters to develop a robust model.

\begin{figure*}[t]
  \centering
  \includegraphics[width=\textwidth]{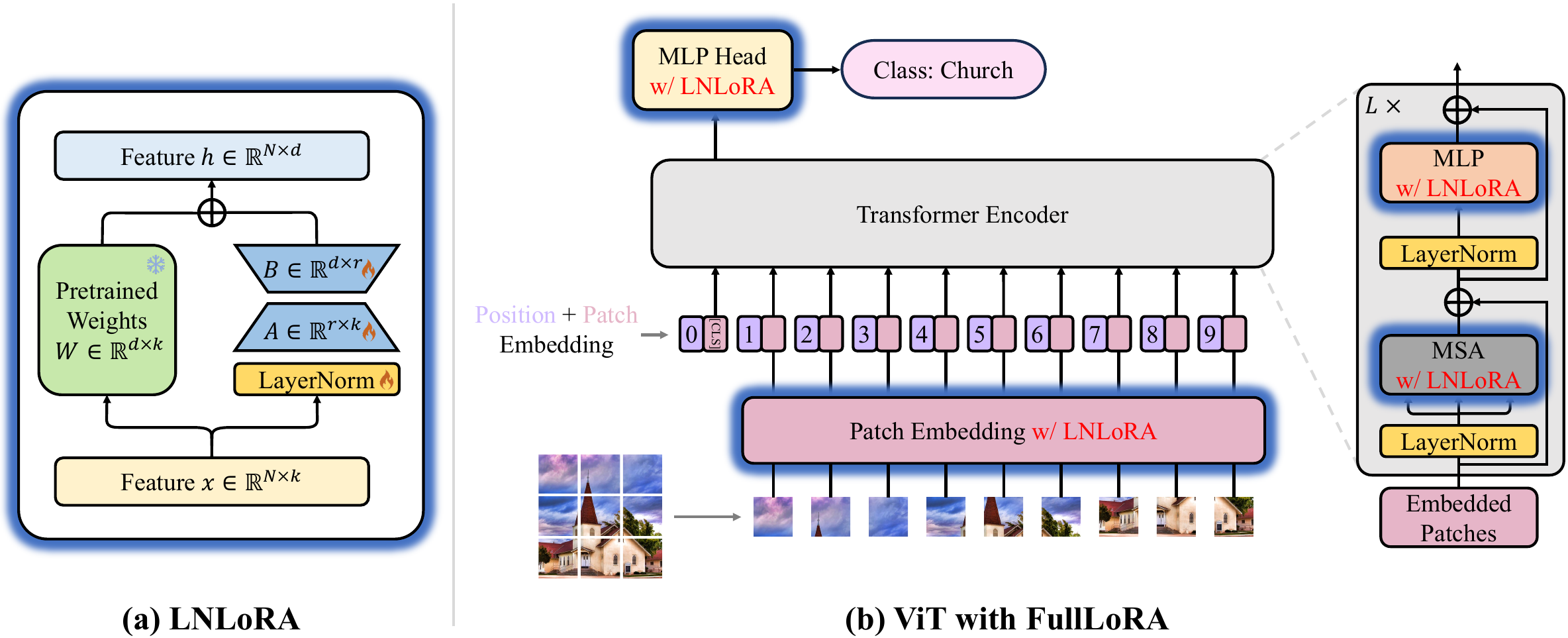}
%   \vspace{-3mm}
  \caption{(a) A diagram of our proposed LNLoRA module. (b) A Vision Transformer (ViT) model utilizing our proposed FullLoRA framework, which employs the LNLoRA module in the patch embedding layer, the multi-headed self-attention (MSA) block, the multi-layer perceptron (MLP) block and the MLP-based classification head. FullLoRA enables the rapid enhancement of the adversarial robustness of the pretrained model through adversarial finetuning with a small number of trainable parameters.}
  \label{fig:framework}
%   \vspace{-5mm}
\end{figure*}

However, when dealing with ViT-based large models, the extensive number of parameters in these models and the immense size of training datasets make it challenging to apply adversarial training due to its considerable computational overhead. Many methods, such as FreeAT~\cite{shafahi2019adversarial} and FastAT~\cite{wong2020fast}, have adopted a single-step generation of adversarial examples in place of the multi-step generation used in traditional adversarial training, significantly reducing the time required for adversarial training. Additionally, works like AGAT~\cite{wu2022towards} have addressed the high computational cost of the attention mechanism in ViT models by adopting an attention-guided partial token discarding method to enhance the efficiency of adversarial training.
Although the aforementioned methods have somewhat reduced the time required for adversarial training, they still necessitate updating all parameters of the model, resulting in significant additional model storage costs, while also incurring notable losses in robustness. Furthermore, in recent years, several methods have been proposed for the lightweight finetuning of pretrained large models in downstream tasks. Parameter-efficient language model tuning methods such as Adapters~\cite{houlsby2019parameter}, Prefix-tuning~\cite{li2020prefix-tuning}, LoRA~\cite{hu2022lora}, UniPELT~\cite{mao2022unipelt}, AdaLoRA~\cite{zhang2023adaptive}, and Compacter~\cite{mahabadi2021compacter} have been introduced for efficient finetuning of large language models (LLMs) in downstream tasks. Recently, Aurora~\cite{wang2023parameter} and VeRA~\cite{kopiczko2024vera} have also been proposed for efficient downstream task finetuning of large-scale multimodal foundation models.
These methods effectively enhance the performance of pretrained models in downstream tasks by finetuning a small number of learnable parameters. However, there has been few work focusing on designing a parameter-efficient approach to improve the adversarial robustness of pretrained models.

To enhance the robustness of models pretrained using standard training in a lightweight manner, we draw inspiration from the LoRA method~\cite{hu2022lora} and introduce the FullLoRA framework. This framework aims to substantially improve the robustness of pretrained models against adversarial examples by adding only a small number of additional learnable parameters, which significantly reduces training time. Specifically, we first analyse the parameters from both adversarially and standardly trained models and reveal notable disparities in the parameters of scale and bias in the layernorm. These disparities can not be adequately addressed by the existing LoRA method, which is demonstrated in \cref{fig:para_diff} of \cref{sec:distance}. To tackle this issue, we introduce the novel LNLoRA module, which integrates a learnable layernorm layer within the existing LoRA module.
Moreover, unlike the existing LoRA method which focuses only on the multi-headed self-attention (MSA) block and multi-layer perceptron (MLP) block in the Transformer layer, our approach introduces LNLoRA modules into more key components of the ViT, \ie, the patch embedding layer, MSA block, MLP block, and the classification head, as shown in~\cref{fig:framework}.
It is worth mentioning that our FullLoRA method is independent of existing advanced adversarial training techniques. This characteristic allows for seamless integration with other methods, leading to further improvement in the efficiency and effectiveness of adversarial finetuning.

% Additionally,  Notably, the structure with the added Pre LN layer can still merge the trainable matrices with the frozen weights when deployed, introducing no inference latency compared to the fully finetuned model.

We evaluate our FullLoRA method on several datasets and ViT-based models. Extensive experiments demonstrate that our proposed FullLoRA method can significantly enhance the robustness of pretrained models against adversarial examples while substantially saving on additional training parameters and accelerating the training speed. In the case of ViT-B on the CIFAR10 dataset, compared to fully finetuned models, our method demonstrates remarkable efficiency, which requires only about 5\% of the learnable parameters to achieve comparable robustness, \ie, incurring a mere 1\%-2\% loss in accuracy under the evaluation of various attacks, \eg, CW~\cite{carlini2017towards}, PGD~\cite{madry2018towards} and AutoAttack~\cite{croce2020reliable}.
% Our research not only underscores the importance of robust model training in the face of adversarial challenges but also opens new avenues for developing lightweight, efficient methods for enhancing the security and reliability of deep learning models in various computer vision tasks.

The primary contributions of our research are summarized as follows:

1. We propose the task of enhancing the adversarial robustness of standardly pretrained models using a lightweight, parameter-efficient approach.

2. We introduce novel LNLoRA module, based on the analysis of the parameters from both adversarially and standardly trained models, to alleviate notable disparities in the parameters of scale and bias in the layernorm.

3. We propose the FullLoRA method, a lightweight parameter-efficient approach to rapidly improve the robustness of the pretrained model, which significantly reduces the number of learnable parameters, thereby substantially saving on model storage space and reducing the time required for model finetuning.

\section{Related Work}
\label{sec:related_work}

%-------------------------------------------------------------------------
\subsection{Vision Transformer}
Vision Transformer (ViT), since its introduction by \cite{dosovitskiy2021an}, has revolutionized the field of computer vision, marking a significant shift from the conventional Convolutional Neural Networks (CNNs)~\cite{he2016deep, szegedy2016rethinking, huang2017densely} that dominated the field for years. ViT employs self-attention mechanisms, typical of transformer models in natural language processing~\cite{vaswani2017attention}, to process images, which distinguishes it from the localized and hierarchical nature of CNNs~\cite{krizhevsky2012imagenet, simonyan2014very}.
In image classification, ViT~\cite{dosovitskiy2021an} has shown remarkable success, achieving state-of-the-art performance on benchmarks like ImageNet~\cite{russakovsky2015imagenet}. Later, various variants of ViT have been proposed to improve the efficiency and performance of ViT models, \ie., DeiT~\cite{touvron2021training}, Swin~\cite{liu2021swin}, MetaFormer~\cite{yu2022metaformer} and PSLT~\cite{wu2023pslt}.
With the great success of ViT in image classification, the application of ViT-based models has also extended to other visual tasks, such as object detection~\cite{carion2020end, fang2021you, wang2021pyramid, wang2022bridged}, semantic segmentation~\cite{zheng2021rethinking, jain2021semask, strudel2021segmenter, cheng2022masked}, image super resolution~\cite{yang2020learning, liang2021swinir, zheng2023efficient}.
In the various visual tasks mentioned above, ViT-based models have achieved significant performance improvements and have gradually become mainstream models in the field of computer vision.

%-------------------------------------------------------------------------
\subsection{Adversarial Training}

Existing research has proposed numerous methods for the adversarial defense of models, such as adversarial training~\cite{madry2018towards}, adversarial purification~\cite{nie2022diffusion}, adversarial example detection~\cite{li2020connecting}, and certifiable defense~\cite{croce2020provable}, with adversarial training being one of the most effective.
Adversarial training is first introduced by \cite{madry2018towards}, which improves model accuracy under adversarial attacks by incorporating generated adversarial examples into the training set. Researches like TRADES~\cite{zhang2019theoretically} and MART~\cite{wang2020improving} further enhance model robustness by optimizing the objective function during adversarial training. RiFT~\cite{zhu2023improving} proposes to exploit the redundant capacity for robustness by fine-tuning the adversarially trained model on its non-robust-critical module to en-hance generalization without compromising adversarial robustness. TWINS~\cite{liu2023twins} explores how to maintain the robustness in the pre-trained model when learning the downstream task, and proposes a novel statistics-based framework, which is effective on a wide range of image classification datasets in terms of both generalization and robustness. PMG-AFT~\cite{wang2024pre} leverages supervi-sion from the original pre-trained model by carefully designing an auxiliary branch, to enhance the model's zero-shot adversarial robustness. Due to the significant time required to generate adversarial examples in adversarial training, methods like FreeAT~\cite{shafahi2019adversarial} and FastAT~\cite{wong2020fast} have markedly accelerated this process through single-step attacks while maintaining high model robustness.
For adversarial training on ViT models, recent studies~\cite{debenedetti2022a, singh2023revisiting} have improved this process from the perspectives of model structure, training strategy, and training tricks. Mo \etal~\cite{mo2022when} improves the adversarial robustness of ViTs by randomly masking gradients from some attention blocks and masking perturbations on some patches during training. AGAT~\cite{wu2022towards} addresses the computational expense of the attention mechanism in ViT models by discarding some tokens guided by attention, thus saving time in adversarial training. Rebuffi \etal~\cite{rebuffi2023revisiting} trains separate classification tokens of a ViT for the clean and adversarial domains, achieving good clean and robust accuracy, which can be further balanced through a model soup approach. AutoLoRA~\cite{xu2023autolora} uses the LoRA~\cite{hu2022lora} mechanism to decouple features of clean and adversarial examples, thereby enhancing the robustness of pretrained models in downstream tasks after finetuning.

Most existing adversarial training methods involve training or finetuning the entire set of model parameters. In contrast, our work freezes most of the pretrained model parameters and proposes to enhance the robustness of pretrained models in a lightweight manner through a small number of additional trainable parameters.

%-------------------------------------------------------------------------
\subsection{Parameter-efficient Model Finetune}

Large models pretrained on extensive datasets exhibit commendable generalizability, making finetuning on such models for downstream tasks a prevalent paradigm. In recent years, numerous parameter-efficient finetuning methods have been proposed for efficiently optimizing models for downstream tasks. In natural language processing (NLP), methods such as Adapters~\cite{houlsby2019parameter}, Prefix-tuning~\cite{li2020prefix-tuning}, LoRA~\cite{hu2022lora}, UniPELT~\cite{mao2022unipelt}, AdaLoRA~\cite{zhang2023adaptive} and Compacter~\cite{mahabadi2021compacter} have been introduced. They add a small number of additional learnable parameters to the pretrained model to facilitate rapid and efficient finetuning for downstream NLP tasks.
% LoRA, targeting large language models (LLMs), injects trainable low-rank matrices into each layer of the Transformer architecture, significantly reducing the number of parameters used during finetuning for downstream tasks. UniPELT combines the lightweight finetuning approaches of LoRA, Adapters, and Prefix-tuning, using a gating mechanism to control the amount of new information introduced by different methods.
In visual tasks, Visual Prompt~\cite{jia2022visual} introduces only a small number of trainable parameters in the input space while keeping the model backbone frozen, often outperforming full finetuning. Furthermore, with the recent rise of multimodal models, many works have focused on the parameter-efficient finetuning of these models. VL-Adapter~\cite{sung2022vl-adapter} introduces adapter-based parameter-efficient transfer learning techniques to vision-language models, and Aurora~\cite{wang2023parameter} applies CP decomposition to model parameters and proposes the informative context enhancement and gated query transformation module for lightweight finetuning. Recently, VeRA~\cite{kopiczko2024vera} proposes to use a single pair of low-rank matrices shared across all layers and learn small scaling vectors instead, which shows the effectiveness both on language and vision tasks.

Existing parameter-efficient model finetuning methods are based on pretrained models and target specific downstream tasks. Our work, however, is the first to propose using a parameter-efficient approach to enhance the robustness of standardly trained models against adversarial examples.
\section{Method}
\label{sec:method}

In this section, we first provide a brief overview of the Vision Transformer~\cite{dosovitskiy2021an} (ViT) model, which has been widely used in various computer vision tasks, as well as the existing Low-Rank Adaptation (LoRA) method~\cite{hu2022lora}, known for its efficiency in lightweight finetuning of pretrained models for various downstream tasks. Following this, we introduce our novel FullLoRA method, a lightweight approach designed to enhance the adversarial robustness of standardly pretrained models in a parameter-efficient manner.

\subsection{Preliminary}
This section begins with a brief overview of the Vision Transformer (ViT) model structure and the existing Low-Rank Adaptation (LoRA) method.

\noindent\textbf{Vision Transformer.}
In recent years, the Vision Transformer (ViT) has achieved remarkable performance improvements in various visual tasks and has gradually become the mainstream model for these tasks.

The ViT model first resizes the input image $x\in \mathbb{R}^{H\times W \times C}$ into different patches $x_p\in\mathbb{R}^{N\times (P^2\cdot C)}$, where $(H, W)$ is the resolution of the image, $C$ is the number of channels in the image, $(P, P)$ is the resolution of each image patch, and $N=HW/P^2$ is the number of patches after resizing. ViT initially maps these image patches to features through a patch embedding layer $E\in \mathbb{R}^{(P^2\cdot C)\times D}$, concatenates them with a special classification token $x_{class}$, and adds position embedding $E_{pos}\in\mathbb{R}^{(N+1)\times D}$ to obtain the embedded patches $z_0$, as:
\begin{equation}
    z_{0} =[x_\text{class};x_p^1E;x_p^2E;\cdots ;x_p^NE]+E_{pos}.
\end{equation}
These embedded patches are then transformed by $L$ consecutive Transformer layers, each consisting of a multi-headed self-attention (MSA) and multi-layer perceptron (MLP) block, resulting in the output $z_L$ of the entire Transformer encoder, which can be formulated as:
\begin{align}
    &z_{l}^{\prime} =\mathrm{MSA}(\mathrm{LN}(z_{l-1}))+z_{l-1},\quad\mathrm{~for~}l=1,\ldots, L  \\
    &z_{l} =\mathrm{MLP}(\mathrm{LN}(z_l^{\prime}))+z_l^{\prime},\quad\mathrm{for~}l=1,\ldots, L
\end{align}
where $\mathrm{LN}$ is the layernorm, $\mathrm{MSA}$ is the multi-headed self-attention block, $\mathrm{MLP}$ is the multi-layer perceptron block.
Finally, the feature $z_L^0$ at the classification token position of $z_L$ is processed through an MLP-based classification head layer to produce the model's final classification result $y$:
\begin{equation}
    y=\mathrm{MLPHead}(z_{L}^{0}),
\end{equation}
where $\mathrm{MLPHead}$ is the MLP-based classification head.

\noindent\textbf{Low-Rank Adaptation.}
Low-rank adaptation (LoRA) is an efficient method for finetuning pretrained models on downstream tasks. It utilizes a low-rank approach for updating additional model weights rather than directly finetuning the pretrained model, effectively reducing the number of parameters and training costs. 

For a pretrained weight matrix $W\in \mathbb{R}^{d\times k}$, LoRA constrains the update during finetuning for downstream tasks by representing the parameter update with a low-rank decomposition: $W + \Delta W = W + BA$, where $B \in \mathbb{R}^{d\times r}$, $A \in \mathbb{R}^{r\times k}$, and the rank $r\leq min(d, k)$.  During the finetuning, the pretrained weight matrix $W$ remains frozen, while $A$ and $B$ serve as the learnable parameters. The LoRA module is typically used in the MSA block and the MLP block of Transformer layers to facilitate parameter-efficient finetuning of pretrained models.

% By this method, large pretrained models can be finetuned more efficiently for specific tasks without the need to retrain or significantly modify the entire model structure. This not only saves computational resources but also allows for retaining much of the useful knowledge and generalization capabilities that the pretrained models have already acquired.

\subsection{FullLoRA Adversarial Training}
Existing Vision Transformer (ViT) models have achieved promising results in various visual tasks. However, during model training, there is often a greater focus on performance, potentially leading to a neglect of the model's robustness. In response to this situation, we propose a new task setting: for pretrained models obtained through standard training, we aim to enhance the adversarial robustness quickly using a parameter-efficient approach for finetuning. To this end, we introduce FullLoRA Adversarial Training, a method which enhances model robustness with minimal additional training parameters. Compared to the commonly used full finetuning approach, our FullLoRA introduces a significantly smaller number of additional trainable parameters, thereby saving on both training time and model storage costs during finetuning.

% Vision Transformer (ViT) models have shown remarkable success across various visual tasks. Nonetheless, a predominant focus on performance during training often sidelines considerations of model robustness. Addressing this gap, we propose a novel paradigm specifically for pretrained models derived from standard training: rapidly boosting the robustness of the model through a parameter-efficient adversarial finetuning approach. To this end, we introduce FullLoRA adversarial training (FullLoRA),
% an innovative method that efficiently enhances the robustness of pretrained models using a minimal amount of additional trainable parameters. Notably, compared to traditional fully finetuning methods, FullLoRA requires significantly fewer additional training parameters, resulting in considerable savings in both model storage requirements and training time during the finetuning process.

In the following, we will provide a detailed introduction to our proposed FullLoRA method.

\begin{figure*}[htbp]
    \centering
    \subfloat[$\|W_{clean}\|$]{
      \includegraphics[width=0.3\textwidth]{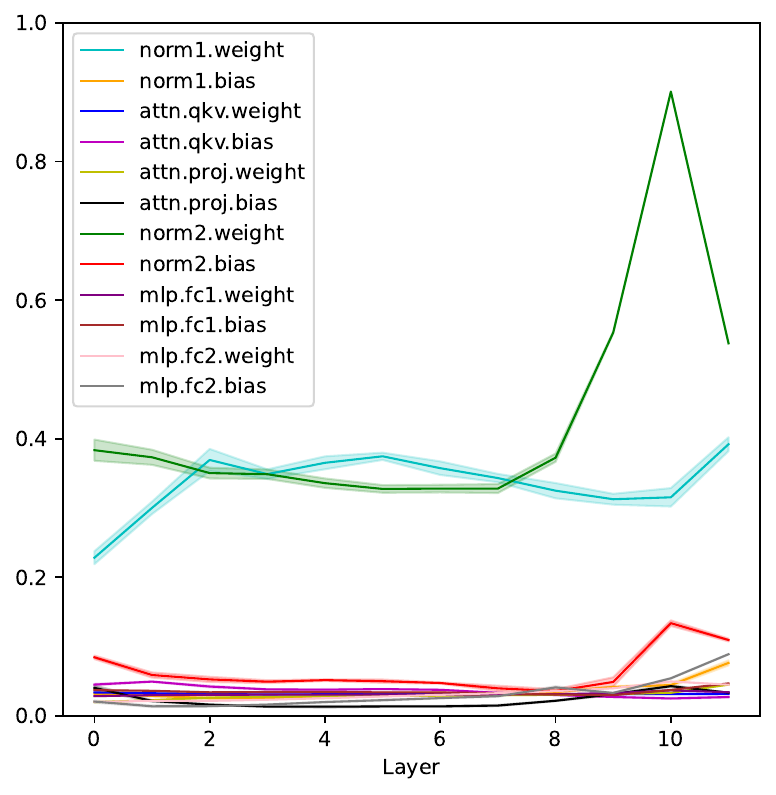}
    }
    \hfil
    \subfloat[$\|W_{adversarial}\|$]{
      \includegraphics[width=0.3\textwidth]{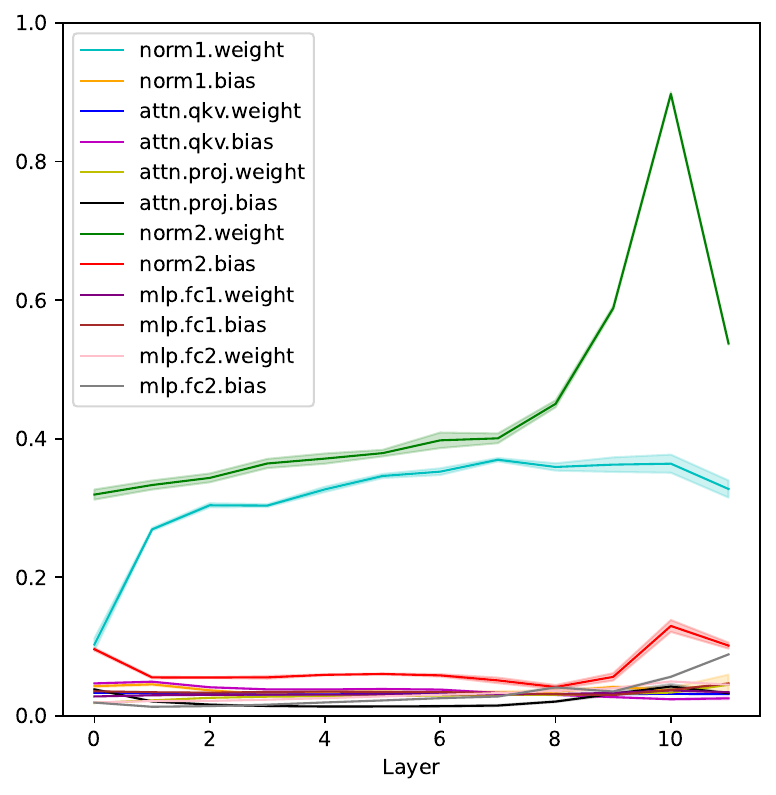}
    }
    \hfil
    \subfloat[$\|W_{clean} - W_{adversarial}\|$]{
      \includegraphics[width=0.3\textwidth]{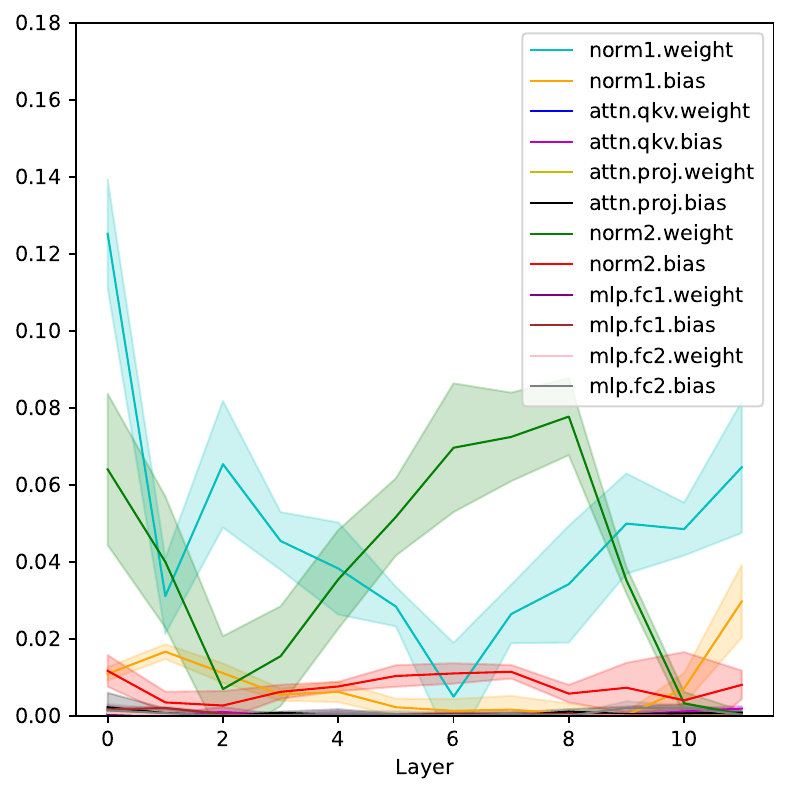}
    }
    \caption{Visualization of the average absolute values of different parameters across the 12 Transformer layers of the ViT model. (a) presents the model with standard training, (b) presents the model with adversarial training, and (c) presents the differences in model parameters between (a) and (b). The experiment is repeated by 5 times, the solid line represents the mean value of the 5 experiments, while the shaded area represents the standard deviation.}
    \label{fig:para}
  \end{figure*}

% \noindent\textbf{LNLoRA.}
\subsubsection{LNLoRA}
We first analyze the differences in parameters between models obtained by standard training and adversarial training. We conduct the experiment on the CIFAR10 dataset~\cite{krizhevsky2009learning} with the ViT-B model~\cite{dosovitskiy2021an}. As shown in~\cref{fig:para}, comparing the average absolute values of different parameters across the 12 Transformer layers, we can clearly see that the scale and bias coefficients in the layernorm preceding each MSA and MLP block in the ViT exhibit particularly significant differences between the two models, especially can be observed from (c) in~\cref{fig:para}, which represents their residuals.

Due to the feature scaling functionality of the layernorm, the features of clean images and adversarial examples in models with standard and adversarial training show significant differences after scaling through the layernorm before entering the MSA and MLP blocks. The existing LoRA method~\cite{hu2022lora} applies only to the weights in the Key, Query, and Value parameters of the MSA block and the two fully connected layers of the MLP block. When freezing the parameters of the standard training model and only applying the existing LoRA method for adversarial finetuning, it becomes challenging for the model to approximate the parameters obtained through fully adversarial training using the low-rank matrices in the LoRA module.

To address these limitations, we introduce the LNLoRA module, which incorporates an additional layernorm preceding the existing LoRA module. The purpose of LNLoRA is to learn the differences in feature magnitude between standard and adversarial trained models via the extra layernorm, allowing the LNLoRA module to better approximate the differences between the parameters of the two models, which can be well validated in~\cref{sec:distance}.
As shown in~\cref{fig:framework}a, consider an image feature represented as $x$, undergoing an initial matrix transformation $h=Wx$.  With the addition of our proposed layernorm-included LNLoRA module, it yields:

\begin{equation}
    h = W x + BA \cdot \textrm{LN}(x) = W x + BA(\alpha \cdot \frac{x-\mu(x)}{\sigma(x)} + \beta),
\end{equation}
 where $\alpha$ and $\beta$ denote the scale and bias coefficients in the layernorm, $\mu(\cdot)$ and $\sigma(\cdot)$ represents the mean and standard deviation of the input. The integration of layernorm into the LNLoRA module significantly enhances the model's ability to adapt to adversarial examples, thereby boosting its robustness with a small number of learnable parameters efficiently. It should be noted that the introduction of layernorm prevents LNLORA from inheriting the LoRA method's characteristic of merging parameters into matrix $W$ during inference. However, as demonstrated in subsequent experiments, the additional computational overhead introduced by this design choice is negligible.

% \noindent\textbf{FullLoRA.}
\subsubsection{FullLoRA}
Based on the aforementioned LNLoRA module, we introduce the FullLoRA method, a novel and lightweight approach designed to enhance the adversarial robustness of standardly pretrained models in a parameter-efficient manner. As depicted in~\cref{fig:framework}b, the LNLoRA module is seamlessly integrated into several key components of the ViT model, including the patch embedding layer,  MSA block, MLP block, and the classification head.

Traditional LoRA methods predominantly focus on modifying parameters within the MSA and MLP blocks of the ViT model, while neglecting the parameters of other model components. However, given the significant differences in image features between clean and adversarial examples, it becomes essential to adaptively finetune the low-level layers of the model, which may facilitate the model's ability to effectively learn features from adversarial examples that are crucial for accurate classification. Therefore, we also encompass the LNLoRA module to the patch embedding layer of the ViT model.

Additionally, given the critical role of parameters in the classification head layer for determining classification results, and the usual practice of unfreezing this layer during finetuning, we also apply the LNLoRA module to the classification head layer. This extension aims to facilitate a more parameter-efficient adaptation of the model when confronting adversarial examples. Our subsequent ablation studies in the experiment further reveal that the integration of LNLoRA across these key components of ViT model, as opposed to the traditional LoRA method, clearly further boosts the robustness of the model against adversarial attacks.

It is worth mentioning that our FullLoRA method is independent of existing advanced adversarial training techniques. This characteristic allows for seamless integration with others, leading to further improvement in the efficiency and effectiveness of adversarial finetuning.

\section{Experiment}
\label{sec:exper}
In this section, we first introduce the experimental setting, including datasets, models, comparison methods, and evaluation metrics. Subsequently, we conduct the ablation study to explore the impact of employing LoRA and LNLoRA modules on different components of the ViT model on model robustness. This is followed by comparisons with existing methods on various models and datasets. We then analyze the impact of using different rank $r$ values on the effectiveness of FullLoRA. Finally, we integrate FullLoRA with other advanced methods to further accelerate the adversarial training and improve the robustness of the model, effectively demonstrating the generalization of our approach.

\subsection{Experimental Setting}

% \noindent\textbf{Dataset.}
\subsubsection{Dataset}
We conduct experiments on four commonly used datasets in the field of adversarial training, \ie, CIFAR10~\cite{krizhevsky2009learning}, CIFAR100~\cite{krizhevsky2009learning}, TinyImageNet~\cite{le2015tiny}, Imagenette~\cite{howard2020fastai} and ImageNet~\cite{russakovsky2015imagenet}. CIFAR10 and CIFAR100 each contains images of 10 and 100 classes respectively, with training and test datasets comprising 50,000 and 10,000 images respectively. The image size in these datasets is 32$\times$32. TinyImageNet contains 100,000 images of 200 classes (500 for each class) downsized to 64$\times$64 colored images. Each class has 500 training images, 50 validation images and 50 test images. Imagenette, a subset of ImageNet~\cite{russakovsky2015imagenet}, includes 10 classes of images, with training and test datasets consisting of 13,000 and 500 images respectively. The image size in TinyImageNet, Imagenette and ImageNet is 224$\times$244.

% \noindent\textbf{Models.}
\subsubsection{Models}
We choose ViT-S~\cite{dosovitskiy2021an}, ViT-B~\cite{dosovitskiy2021an} and Swin-B~\cite{liu2021swin} models to conduct the experiments. In all experiments, we initialize the model with weights pretrained on ImageNet-1K while employing random initialization for the final classification layer.

% \noindent\textbf{Comparison methods.}
\subsubsection{Comparison Methods}
In our study, we compare our FullLoRA method with other approaches such as LoRA~\cite{hu2022lora}, UniPELT~\cite{mao2022unipelt}, and Aurora~\cite{wang2023parameter}, which are all methods for parameter-efficient finetuning on pretrained models. Additionally, following~\cite{rebuffi2023revisiting}, we also compare our approach against strategies that exclusively update specific components during finetuning, \eg, the patch embedding layer and the scale and offset parameters of all layernorms. It should be noted that, in our comparisons, we have combined these methods with adversarial training to enhance the model's robustness, ensuring a fair comparison.
%This comparative analysis will provide a comprehensive understanding of the efficacy and efficiency of our FullLoRA method in enhancing model robustness against adversarial samples, relative to other parameter-efficient finetuning methods.

% \noindent\textbf{Evaluation metrics.}
\subsubsection{Evaluation Metrics}
In our evaluation of the robustness of various models against adversarial examples, we employ accuracy metrics under different adversarial attack methods. Specifically, we utilize CW-20~\cite{carlini2017towards}, PGD-20, PGD-100~\cite{madry2018towards}, and AutoAttack~\cite{croce2020reliable} to evaluate model robustness. CW-$k$ and PGD-$k$ denote CW and PGD attacks with $k$ iterative steps, respectively. The perturbation magnitude for these attacks is set to 8/255, with a step size of 1/255. Additionally, we evaluate the efficiency of finetuning pretrained models for robustness by considering the number of learnable parameters required during finetuning and the time taken for the finetuning process.

% \noindent\textbf{Details.}
\subsubsection{Details}
In our experiments, finetuning on all datasets is conducted over 40 epochs, with the learning rate being scaled down by a factor of 0.1 after the 35-th and 38-th epochs. In default, we employ standard adversarial training with a PGD-10 attack as our baseline method. In our adversarial training, we exclusively use adversarial examples and does not utilize clean images. The SGD optimizer~\cite{robbins1951stochastic} is utilized for the training process. The learning rate settings vary depending on the finetuning method employed: for fully finetuned models and Aurora~\cite{wang2023parameter}, a learning rate of 0.1 is used; for LoRA~\cite{hu2022lora} and our FullLoRA, the learning rate is set to 5; for UniPELT~\cite{mao2022unipelt}, the learning rate is 0.5; and for the remaining methods, a learning rate of 1 is applied. In the case of LoRA and FullLoRA, considering the trade-off between model performance and training efficiency, we set the rank $r$ to 32 by default. The experiments are carried out on a single NVIDIA GeForce RTX 3090 except ImageNet, which is conducted on four NVIDIA GeForce RTX 3090. In order to ensure the stability and persuasiveness of the experiment, we re-conduct most experiments in our proposed FullLoRA method five times, and show the mean and standard deviation of experimental results.

\begin{table*}[htbp]
    \centering
    \caption{Ablation study on the use of LoRA/LNLoRA in different components of the ViT model.}
    % \vspace{-3mm}
    \resizebox{\textwidth}{!}{
        \begin{tabular}{c|c|c|c|c|c|c|ccccc|c}
            \hline
            \multirow{2}[4]{*}{Method} & \multicolumn{5}{c|}{Setting}          & Trainable & Clean Acc & CW-20 & PGD-20 & PGD-100 & AutoAttack & Time \\
            \cline{2-6}      & Patch Embedding & MSA block & MLP block & Classification Head & LNLoRA & Paras (M) & (\%)  & (\%)  & (\%)  & (\%)  & (\%)  & (h) \\
            \hline
            Fully Finetune & ---   & ---   & ---   & ---   & ---   & 85.15  & 82.83  & 50.26  & 52.85  & 52.56  & 48.26  & 11.17  \\
            \hline
            LoRA~\cite{hu2022lora} &       & $\surd$ & $\surd$ &       & w/o   & 4.72  & 84.11  & 47.14  & 49.96  & 49.74  & 45.13  & 8.10  \\
            (a)   &       & $\surd$ & $\surd$ & $\surd$ & w/o   & 4.74  & 84.36  & 47.53  & 50.31  & 50.20  & 45.59  & 8.17  \\
            (b)   & $\surd$ & $\surd$ & $\surd$ &       & w/o   & 4.74  & 84.21  & 47.68  & 50.49  & 50.31  & 45.67  & 8.18  \\
            (c)   & $\surd$ & $\surd$ & $\surd$ & $\surd$ & w/o   & 4.77  & 85.55  & 48.75  & 50.83  & 50.43  & 46.26  & 8.23  \\
            \hline
            FullLoRA (ours) & $\surd$ & $\surd$ & $\surd$ & $\surd$ & w/  & 4.85  & 87.12  & 49.76  & 51.38  & 50.83  & 46.94  & 8.38  \\
            \hline
        \end{tabular}%
    }
    \label{tab:ablation_study}%
    % \vspace{-5mm}
\end{table*}%

\subsection{Ablation Study}
We conduct the ablation study using LoRA and LNLoRA modules on different components of the ViT model, and the results are presented in~\cref{tab:ablation_study}. The experiment is conducted using the ViT-B model on the CIFAR10 dataset. As clearly shown, compared to the original LoRA method which only adds low-rank matrices in the MSA block and MLP block for finetuning, the application of the LoRA module additionally in the classification head and patch embedding layer respectively (\ie, (a) and (b) in~\cref{tab:ablation_study}) improves the robustness of model under various attacks. Employing the LoRA module across all four mentioned components (\ie, (c) in~\cref{tab:ablation_study}) further enhances the robustness of the model. Moreover, replacing the LoRA module with our proposed LNLoRA module can further improve the robustness, demonstrating the greater suitability of our newly proposed LNLoRA module for adversarial finetuning on pretrained models. Additionally, as different modules are incrementally added, not only does the robustness of the model improve, but also the clean accuracy of the model increases consistently, which further demonstrates the effectiveness of our proposed FullLoRA.

\begin{table*}[htbp]
    \centering
    % \caption{Comparison with other methods across different datasets. The results are obtained using the ViT-B model.}
    \caption{Comparison with other methods across different datasets. The results are obtained using the ViT-B model. The results reported for the FullLoRA method are the mean and standard deviation of repeating the experiment five times.}
    % \vspace{-3mm}
    % \resizebox{\columnwidth}{!}{
        \begin{tabular}{c|c|c|ccccc|c}
            \hline
            \multirow{2}[2]{*}{Dataset} & \multirow{2}[2]{*}{Method} & Trainable & Clean Acc & CW-20 & PGD-20 & PGD-100 & AutoAttack & Time \\
                  &       & Paras (M) & (\%)  & (\%)  & (\%)  & (\%)  & (\%)  & (h) \\
            \hline
            \multirow{8}[4]{*}{CIFAR-10} & Standard Training & ---   & 98.65  & 0.00  & 0.00  & 0.00  & 0.00  & --- \\
                  & Fully Finetune & 85.15 & 82.83  & 50.26  & 52.85  & 52.56  & 48.26  & 11.17  \\
            \cline{2-9}      & Patch Embedding Layer & 0.04  & 41.14  & 15.77  & 19.91  & 17.72  & 15.13  & 7.42  \\
                  & Layer Normalization & 0.04  & 35.64  & 20.87  & 23.58  & 23.57  & 20.97  & 7.79  \\
                  & Aurora~\cite{wang2023parameter} & 0.09  & 76.09  & 38.50  & 41.85  & 41.67  & 39.87  & 11.48  \\
                  & LoRA~\cite{hu2022lora} & 4.72  & 84.11  & 47.14  & 49.96  & 49.74  & 45.13  & 8.10  \\
                  & UniPELT~\cite{mao2022unipelt} & 11.42 & 84.63  & 47.63  & 49.49  & 48.99  & 44.63  & 13.11  \\
                  & FullLoRA (ours) & 4.85  & \textbf{87.16$\pm$0.20} & \textbf{50.36$\pm$0.12} & \textbf{51.50$\pm$0.09} & \textbf{51.04$\pm$0.08} & \textbf{46.97$\pm$0.09} & 8.39$\pm$0.02 \\
            \hline
            \multirow{8}[4]{*}{Imagenette} & Standard Training & ---   & 98.00  & 0.00  & 0.20  & 0.00  & 0.00  & --- \\
                  & Fully Finetune & 85.15 & 93.40  & 69.00  & 68.80  & 68.60  & 67.60  & 8.52  \\
            \cline{2-9}      & Patch Embedding Layer & 0.04  & 37.60  & 23.40  & 25.60  & 25.60  & 24.80  & 5.76  \\
                  & Layer Normalization & 0.04  & 61.40  & 23.60  & 28.80  & 28.80  & 27.60  & 5.89  \\
                  & Aurora~\cite{wang2023parameter} & 0.09  & 87.20  & 51.40  & 53.20  & 53.00  & 51.80  & 9.07  \\
                  & LoRA~\cite{hu2022lora} & 4.72  & \textbf{94.00} & 63.80  & 62.00  & 61.20  & 60.00  & 6.15  \\
                  & UniPELT~\cite{mao2022unipelt} & 11.42 & 91.40  & 58.00  & 57.80  & 56.20  & 55.40  & 9.58  \\
                  & FullLoRA (ours) & 4.85  & 93.48$\pm$0.23 & \textbf{66.26$\pm$0.26} & \textbf{66.28$\pm$0.23} & \textbf{65.64$\pm$0.17} & \textbf{64.02$\pm$0.11} & 6.31$\pm$0.02 \\
            \hline
        \end{tabular}%
    % }
    \label{tab:comparison_dataset}%
    % \vspace{-3mm}
\end{table*}%

\subsection{Comparison with Other Methods}
We first compare the robustness against different attack methods of pretrained models finetuned using our proposed FullLoRA method with other lightweight finetuning methods on various datasets. It should be noted that, in our comparisons, we have combined these methods with adversarial training to enhance the model's robustness, ensuring a fair comparison. The results are shown in~\cref{tab:comparison_dataset}. When only the patch embedding layer or the scale and offset parameters of layer normalization are used for adversarial finetuning, the number of trainable parameters is minimal and the training speed is relatively fast, whereas the robustness of the model in this scenario is weak. While Aurora~\cite{wang2023parameter}, LoRA~\cite{hu2022lora}, and UniPELT~\cite{mao2022unipelt} methods show improvements in model robustness, the Aurora method involves a large number of additional matrix multiplication operations due to CP decomposition of the weight matrices, leading to a longer training time, even exceeding that of full finetuning. Also, the UniPELT method adds a considerable amount of extra parameters, resulting in unsatisfactory training duration.

Our proposed FullLoRA method shows a robustness improvement of 2\% $\sim$ 4\% on average across different datasets compared to the LoRA method, with similar parameter count and training duration, which clearly demonstrates the effectiveness of our approach.
% When the rank is decreased to $r = 16$, our FullLoRA can still achieve similar or even better model robustness compared to the LoRA method while requiring only half the number of parameters for finetuning and reducing training time.
Moreover, compared to fully finetuning the entire model, our method uses only about 5\% of the parameters and 75\% of the training duration to achieve comparable model robustness. 
On the dataset of CIFAR-10, our method incurs an average loss of only about 1\% in robust accuracy, yet it simultaneously achieves a 4.29\% improvement in clean accuracy.% More experiments on CIFAR-100 and TinyImageNet are provided in~\cref{app:dataset}.

We also conduct experiments on different ViT-based models using the CIFAR10 dataset, and the results are displayed in~\cref{tab:comparison_model}. Similarly, our method achieves the best results across different models, demonstrating the superiority and generalization of our proposed FullLoRA method.
To further demonstrate the effectiveness of our FullLoRA method, we compare the robustness of different models under the gradient-free attacks (\ie, ZOO~\cite{chen2017zoo}) and obfuscated gradient attacks (\ie, BPDA~\cite{athalye2018obfuscated}). From \cref{tab:comparison_model}, we can clearly see that our proposed FullLoRA method also demonstrates better robustness under the ZOO and BPDA attacks compared to existing methods. This strongly demonstrates the generalization of FullLoRA, showing that it improves the model's robustness under various attack methods.

\begin{table*}[thbp]
    \centering
    % \caption{Comparison with other methods on different ViT-based models. The results are obtained using the CIFAR10 dataset.}
    \caption{Comparison with other methods on different ViT-based models. The results are obtained using the CIFAR10 dataset. The results reported for the FullLoRA method are the mean and standard deviation of repeating the experiment five times.}
    % \vspace{-3mm}
    \resizebox{\textwidth}{!}{
        \begin{tabular}{c|c|c|ccccccc|c}
            \hline
            \multirow{2}[2]{*}{Model} & \multirow{2}[2]{*}{Method} & Trainable & Clean Acc & CW-20 & PGD-20 & PGD-100 & AutoAttack & BPDA  & ZOO   & Time \\
                  &       & Paras (M) & (\%)  & (\%)  & (\%)  & (\%)  & (\%)  & (\%)  & (\%)  & (h) \\
            \hline
            \multirow{6}[4]{*}{ViT-B} & Standard Training & ---   & 98.65  & 0.00  & 0.00  & 0.00  & 0.00  & 0.03  & 1.32  & --- \\
                  & Fully Finetune & 85.15 & 82.83  & 50.26  & 52.85  & 52.56  & 48.26  & 49.63  & 61.38  & 11.17  \\
            \cline{2-11}      & Aurora~\cite{wang2023parameter} & 0.09  & 76.09  & 38.50  & 41.85  & 41.67  & 39.87  & 37.21  & 49.52  & 11.48  \\
                  & LoRA~\cite{hu2022lora} & 4.72  & 84.11  & 47.14  & 49.96  & 49.74  & 45.13  & 45.71  & 56.79  & 8.10  \\
                  & UniPELT~\cite{mao2022unipelt} & 11.42 & 84.63  & 47.63  & 49.49  & 48.99  & 44.63  & 45.32  & 55.17  & 13.11  \\
                  & FullLoRA (ours) & 4.85  & \textbf{87.16$\pm$0.20} & \textbf{50.36$\pm$0.12} & \textbf{51.50$\pm$0.09} & \textbf{51.04$\pm$0.08} & \textbf{46.97$\pm$0.09} & \textbf{47.65$\pm$0.27} & \textbf{59.46$\pm$0.33} & 8.39$\pm$0.02 \\
            \hline
            \multirow{6}[4]{*}{ViT-S} & Standard Training & ---   & 96.09  & 0.00  & 0.00  & 0.00  & 0.00  & 0.03  & 1.74  & --- \\
                  & Fully Finetune & 47.34 & 79.94  & 48.02  & 50.84  & 50.62  & 46.22  & 46.92  & 58.34  & 5.31  \\
            \cline{2-11}      & Aurora~\cite{wang2023parameter} & 0.08  & 70.52  & 36.12  & 38.91  & 38.79  & 35.24  & 36.24  & 44.87  & 5.46  \\
                  & LoRA~\cite{hu2022lora} & 3.15  & 78.31  & 44.87  & 46.10  & 45.98  & 41.37  & 42.67  & 52.68  & 3.87  \\
                  & UniPELT~\cite{mao2022unipelt} & 7.75  & 73.40  & 39.82  & 43.68  & 43.49  & 40.09  & 41.38  & 50.96  & 5.93  \\
                  & FullLoRA (ours) & 3.25  & \textbf{78.97$\pm$0.17} & \textbf{45.69$\pm$0.10} & \textbf{48.57$\pm$0.10} & \textbf{48.41$\pm$0.07} & \textbf{44.95$\pm$0.06} & \textbf{45.19$\pm$0.31} & \textbf{55.83$\pm$0.29} & 3.92$\pm$0.01 \\
            \hline
            \multirow{6}[4]{*}{Swin-B} & Standard Training & ---   & 97.88  & 0.00  & 0.00  & 0.00  & 0.00  & 0.02  & 1.16  & --- \\
                  & Fully Finetune & 86.69 & 81.29  & 48.36  & 50.99  & 50.88  & 46.57  & 46.98  & 58.76  & 6.44  \\
            \cline{2-11}      & Aurora~\cite{wang2023parameter} & 0.09  & 75.13  & 37.52  & 38.17  & 38.03  & 35.26  & 36.61  & 45.13  & 4.79  \\
                  & LoRA~\cite{hu2022lora} & 6.16  & 80.94  & 45.32  & 46.50  & 46.37  & 43.08  & 43.84  & 54.62  & 4.66  \\
                  & UniPELT~\cite{mao2022unipelt} & 11.58 & 79.55  & 44.16  & 45.28  & 44.93  & 41.39  & 42.42  & 53.55  & 5.07  \\
                  & FullLoRA (ours) & 6.29  & \textbf{83.66$\pm$0.19} & \textbf{46.78$\pm$0.11} & \textbf{48.23$\pm$0.11} & \textbf{48.13$\pm$0.12} & \textbf{44.21$\pm$0.09} & \textbf{45.91$\pm$0.21} & \textbf{56.97$\pm$0.17} & 4.81$\pm$0.01 \\
            \hline
        \end{tabular}%
    }
    \label{tab:comparison_model}%
    % \vspace{-3mm}
\end{table*}%

\subsection{Experiments on More Datasets}
\label{app:dataset}
  We further conduct experiments on the CIFAR-100, TinyImageNet, and ImageNet datasets. The results in~\cref{tab:more_dataset} clearly show that our method also achieves the best results on these three datasets, fully demonstrating the good generalizability of the approach we propose. On the CIFAR-100 and TinyImageNet datasets, our proposed FullLoRA method has shown an average robustness improvement of over 3\% compared to LoRA under various attacks, effectively showcasing the superiority of our method. On the more challenging ImageNet dataset, due to training costs and time constraints, we only compare the results with the LoRA method. From the table, it can be seen that our FullLoRA also achieves an average robustness improvement of over 2\% compared to the LoRA method.

\begin{table*}[thbp]
  \centering
%   \caption{More results on the dataset of CIFAR-100, TinyImageNet and ImageNet.}
  \caption{More results on the dataset of CIFAR-100, TinyImageNet and ImageNet. The results reported for the FullLoRA method are the mean and standard deviation of repeating the experiment five times.}
%   \resizebox{\columnwidth}{!}{
    \begin{tabular}{c|c|c|ccccc|c}
        \hline
        \multirow{2}[2]{*}{Dataset} & \multirow{2}[2]{*}{Method} & Trainable & Clean Acc & CW-20 & PGD-20 & PGD-100 & AutoAttack & Time \\
              &       & Paras (M) & (\%)  & (\%)  & (\%)  & (\%)  & (\%)  & (h) \\
        \hline
        \multirow{6}[4]{*}{CIFAR-100} & Standard Training & ---   & 78.39 & 0.00  & 0.00  & 0.00  & 0.00  & --- \\
              & Fully Finetune & 85.22 & 61.02  & 28.61  & 30.58  & 30.43  & 27.49  & 11.18  \\
        \cline{2-9}      & Aurora~\cite{wang2023parameter} & 0.09  & 55.17  & 19.27  & 19.81  & 19.73  & 17.52  & 11.48  \\
              & LoRA~\cite{hu2022lora} & 4.72  & \textbf{63.91} & 24.58  & 25.83  & 25.71  & 22.53  & 8.11  \\
              & UniPELT~\cite{mao2022unipelt} & 11.42 & 55.38  & 21.63  & 24.69  & 24.55  & 22.07  & 13.11  \\
              & FullLoRA (ours) & 4.85  & 61.23$\pm$0.12 & \textbf{25.86$\pm$0.05} & \textbf{28.40$\pm$0.04} & \textbf{28.21$\pm$0.03} & \textbf{25.80$\pm$0.13} & 8.38$\pm$0.02 \\
        \hline
        \multirow{6}[4]{*}{TinyImageNet} & Standard Training & ---   & 84.89  & 0.00  & 0.00  & 0.00  & 0.00  & --- \\
              & Fully Finetune & 85.37  & 44.79  & 24.47  & 23.46  & 22.29  & 22.07  & 22.96  \\
        \cline{2-9}      & Aurora~\cite{wang2023parameter} & 0.09  & 36.49  & 18.23  & 17.92  & 17.37  & 17.06  & 23.49  \\
              & LoRA~\cite{hu2022lora} & 4.72  & 45.12  & 21.37  & 19.79  & 19.01  & 18.42  & 16.29  \\
              & UniPELT~\cite{mao2022unipelt} & 11.42  & 43.17  & 20.56  & 18.14  & 17.62  & 17.04  & 25.57  \\
              & FullLoRA (ours) & 4.87  & \textbf{45.25$\pm$0.17} & \textbf{23.73$\pm$0.31} & \textbf{22.22$\pm$0.18} & \textbf{21.09$\pm$0.11} & \textbf{20.80$\pm$0.09} & 16.59$\pm$0.03 \\
        \hline
        \multirow{4}[4]{*}{ImageNet} & Standard Training & ---   & 85.79  & 0.00  & 0.00  & 0.00  & 0.00  & --- \\
              & Fully Finetune & 86.79  & 73.82  & 52.96  & 51.73  & 48.29  & 47.39  & 207  \\
        \cline{2-9}      & LoRA~\cite{hu2022lora} & 4.86  & 73.84  & 47.12  & 45.61  & 44.85  & 42.32  & 151  \\
              & FullLoRA (ours) & 4.98  & \textbf{73.98} & \textbf{49.37} & \textbf{48.04} & \textbf{46.78} & \textbf{45.17} & 153  \\
        \hline
    \end{tabular}%
%   }
  \label{tab:more_dataset}%
\end{table*}%

\subsection{Results under Different Rank $r$}

\begin{table*}[thbp]
    \centering
    \caption{Results of our FullLoRA under different rank $r$.}
    % \vspace{-3mm}
    % \resizebox{\columnwidth}{!}{
        \begin{tabular}{c|c|c|ccccc|c}
            \hline
            \multirow{2}[2]{*}{Method} & \multirow{2}[2]{*}{Rank $r$} & Trainable & Clean Acc & CW-20 & PGD-20 & PGD-100 & AutoAttack & Time \\
                &       & Paras (M) & (\%)  & (\%)  & (\%)  & (\%)  & (\%)  & (h) \\
            \hline
            Fully Finetune & ---   & 85.15  & 82.83  & 50.26  & 52.85  & 52.56  & 48.26  & 11.17  \\
            \hline
            \multirow{2}[8]{*}{FullLoRA (ours)} & 8     & 1.27  & 84.52  & 46.60  & 49.26  & 48.91  & 45.10  & 7.87 \\
            \cline{2-9}      & 16    & 2.46  & 85.93  & 47.68  & 50.33  & 49.46  & 45.82  & 8.01 \\
            \cline{2-9}      & 32    & 4.85  & 87.12  & 49.76  & 51.38  & 50.83  & 46.94  & 8.38 \\
            \cline{2-9}      & 64    & 9.62  & 86.82  & 50.25  & 51.50  & 51.19  & 47.12  & 8.72 \\
            \hline
        \end{tabular}%
    % }
    \label{tab:rank}%
    % \vspace{-3mm}
\end{table*}%
We compare the robustness of pretrained models adversarially finetuned using our proposed FullLoRA method with different ranks $r$ in the LNLoRA module. The experiment is conducted using the ViT-B model on the CIFAR10 dataset. As shown in~\cref{tab:rank}, as $r$ increases from 8 to 64, clean accuracy improves from 84.52\% to 86.82\%, and the robustness against various attack methods is also enhanced, with an improvement all over 2\%. When comparing the results at $r=64$ with the fully finetuned model, our FullLoRA method utilizes only 11.3\% of the training parameters while achieves comparable results and significantly reduces the training time.
Considering the trade-off between the number of parameters used for adversarial finetuning and the robustness of the model, we choose rank $r=32$ as the default for other experiments in our work.

\begin{table*}[t]
    \centering
    \caption{Results of combining FullLoRA with other speed-accelerating adversarial training method AGAT.}
    % \vspace{-3mm}
    % \resizebox{\columnwidth}{!}{
        \begin{tabular}{c|c|c|ccccc|c}
            \hline
            \multirow{2}[2]{*}{Dataset} & \multirow{2}[2]{*}{Method} & Trainable & Clean Acc & CW-20 & PGD-20 & PGD-100 & AutoAttack & Time \\
                &       & Paras (M) & (\%)  & (\%)  & (\%)  & (\%)  & (\%) & (h) \\
            \hline
            \multirow{2}[2]{*}{CIFAR-10} & AGAT~\cite{wu2022towards} & 85.15  & \textbf{87.42} & 45.02  & 47.75  & 47.54  & 42.86 & 5.76 \\
                & FullLoRA (ours) & 4.85  & 87.12  & \textbf{49.76} & \textbf{51.38} & \textbf{50.83} & \textbf{46.94} & 8.38  \\
            \hline
            \multirow{2}[2]{*}{CIFAR-100} & AGAT~\cite{wu2022towards} & 85.22  & \textbf{63.91} & 25.36  & 25.74  & 25.61  & 24.37 & 5.77   \\
                & FullLoRA (ours) & 4.85  & 61.20  & \textbf{25.86} & \textbf{28.40} & \textbf{28.23} & \textbf{25.78} & 8.38 \\
            \hline
            \multirow{2}[2]{*}{Imagenette} & AGAT~\cite{wu2022towards} & 85.15  & \textbf{94.70} & 61.90  & 61.60  & 61.10  & 60.60 & 4.15   \\
                & FullLoRA (ours) & 4.85  & 93.40  & \textbf{66.20} & \textbf{66.20} & \textbf{65.60} & \textbf{64.00} & 6.32 \\
            \hline
        \end{tabular}%
    % }
    \label{tab:combine_agat}%
    % \vspace{-5mm}
\end{table*}%

\begin{table*}[t]
    \centering
    \caption{Results of combining FullLoRA with other speed-accelerating adversarial training methods.}
    % \vspace{-3mm}
    % \resizebox{\columnwidth}{!}{
        \begin{tabular}{c|c|c|ccccc|c}
            \hline
            \multirow{2}[2]{*}{Dataset} & \multirow{2}[2]{*}{Method} & Trainable & Clean Acc & CW-20 & PGD-20 & PGD-100 & AutoAttack & Time \\
                &       & Paras (M) & (\%)  & (\%)  & (\%)  & (\%)  & (\%)  & (h) \\
            \hline
            \multirow{2}[2]{*}{CIFAR-10} & AGAT~\cite{wu2022towards}+LoRA~\cite{hu2022lora} & 4.72  & 86.34  & 42.68  & 43.89  & 43.65  & 39.68  & 3.81  \\
                & AGAT~\cite{wu2022towards}+FullLoRA (ours) & 4.85  & \textbf{89.47} & \textbf{44.07} & \textbf{45.00} & \textbf{44.45} & \textbf{41.03} & 3.93  \\
            \hline
            \multirow{2}[2]{*}{CIFAR-100} & AGAT~\cite{wu2022towards}+LoRA~\cite{hu2022lora} & 4.72  & 65.18  & 21.73  & 21.99  & 21.86  & 20.42  & 3.82  \\
                & AGAT~\cite{wu2022towards}+FullLoRA (ours) & 4.85  & \textbf{66.31} & \textbf{22.93} & \textbf{23.07} & \textbf{23.01} & \textbf{22.54} & 3.93  \\
            \hline
            \multirow{2}[2]{*}{Imagenette} & AGAT~\cite{wu2022towards}+LoRA~\cite{hu2022lora} & 4.72  & 93.00  & 57.20  & 56.20  & 56.00  & 55.10  & 3.19  \\
                & AGAT~\cite{wu2022towards}+FullLoRA (ours) & 4.85  & \textbf{95.20} & \textbf{60.20} & \textbf{60.00} & \textbf{59.70} & \textbf{58.90} & 3.23  \\
            \hline
        \end{tabular}%
    % }
    \label{tab:combine_with_agat}%
    % \vspace{-5mm}
\end{table*}%

\begin{table*}[thbp]
    \centering
    \caption{Results of combining FullLoRA with other robustness-enhancing adversarial training methods.}
    % \vspace{-3mm}
    % \resizebox{\columnwidth}{!}{
        \begin{tabular}{c|c|c|ccccc|c}
            \hline
            \multirow{2}[2]{*}{Dataset} & \multirow{2}[2]{*}{Method} & Trainable & Clean Acc & CW-20 & PGD-20 & PGD-100 & AutoAttack & Time \\
                &       & Paras (M) & (\%)  & (\%)  & (\%)  & (\%)  & (\%)  & (h) \\
            \hline
            \multirow{4}[8]{*}{CIFAR-10} & MART~\cite{wang2020improving} & 85.15  & 83.65  & 51.09  & 53.62  & 53.41  & 49.02  & 12.13  \\
            \cline{2-9}      & MART~\cite{wang2020improving}+LoRA~\cite{hu2022lora} & 4.72  & 84.82  & 47.93  & 50.77  & 50.55  & 46.01  & 8.86  \\
                & MART~\cite{wang2020improving}+FullLoRA (ours) & 4.85  & \textbf{87.99} & \textbf{50.59} & \textbf{52.09} & \textbf{51.64} & \textbf{47.66} & 8.97  \\
            \cline{2-9}      & PRM~\cite{mo2022when} & 85.15  & 83.43  & 50.97  & 53.49  & 53.12  & 48.81  & 11.23  \\
            \cline{2-9}      & PRM~\cite{mo2022when}+LoRA~\cite{hu2022lora} & 4.72  & 84.62  & 47.71  & 50.53  & 50.45  & 45.78  & 8.16  \\
                & PRM~\cite{mo2022when}+FullLoRA (ours) & 4.85  & \textbf{87.76} & \textbf{50.39} & \textbf{51.82} & \textbf{51.42} & \textbf{47.36} & 8.39  \\
            \hline
            \multirow{4}[8]{*}{CIFAR-100} & MART~\cite{wang2020improving} & 85.22  & 61.82  & 29.51  & 31.46  & 31.20  & 28.23  & 12.15  \\
            \cline{2-9}      & MART~\cite{wang2020improving}+LoRA~\cite{hu2022lora} & 4.72  & 61.70  & 25.44  & 26.59  & 26.60  & 23.36  & 8.88  \\
                & MART~\cite{wang2020improving}+FullLoRA (ours) & 4.85  & \textbf{62.00} & \textbf{26.60} & \textbf{29.22} & \textbf{29.13} & \textbf{26.68} & 8.99  \\
            \cline{2-9}      & PRM~\cite{mo2022when} & 85.22  & 61.46  & 29.15  & 31.12  & 31.12  & 28.13  & 11.25  \\
            \cline{2-9}      & PRM~\cite{mo2022when}+LoRA~\cite{hu2022lora} & 4.72  & \textbf{62.59} & 25.11  & 26.26  & 26.17  & 22.97  & 8.18  \\
                & PRM~\cite{mo2022when}+FullLoRA (ours) & 4.85  & 61.81  & \textbf{26.35} & \textbf{28.88} & \textbf{28.70} & \textbf{26.35} & 8.40  \\
            \hline
            \multirow{4}[8]{*}{Imagenette} & MART~\cite{wang2020improving} & 85.15  & 94.26  & 69.81  & 69.68  & 69.40  & 68.49  & 9.32  \\
            \cline{2-9}      & MART~\cite{wang2020improving}+LoRA~\cite{hu2022lora} & 4.72  & 94.06  & 64.57  & 62.78  & 62.00  & 60.88  & 6.75  \\
                & MART~\cite{wang2020improving}+FullLoRA (ours) & 4.85  & \textbf{94.19} & \textbf{66.99} & \textbf{67.10} & \textbf{66.51} & \textbf{64.91} & 6.83  \\
            \cline{2-9}      & PRM~\cite{mo2022when} & 85.15  & 93.95  & 69.68  & 69.47  & 69.07  & 68.22  & 8.59  \\
            \cline{2-9}      & PRM~\cite{mo2022when}+LoRA~\cite{hu2022lora} & 4.72  & \textbf{94.07} & 64.48  & 62.54  & 61.63  & 60.49  & 6.19  \\
                & PRM~\cite{mo2022when}+FullLoRA (ours) & 4.85  & 94.04  & \textbf{66.69} & \textbf{66.83} & \textbf{66.22} & \textbf{64.64} & 6.35  \\
            \hline
        \end{tabular}%
    % }
    \label{tab:combine_performance}%
    % \vspace{-5mm}
\end{table*}%

\begin{figure*}[htbp]
    \centering
    \subfloat[$\|W_{lora} - W_{adversarial}\|$]{
      \includegraphics[width=0.41\textwidth]{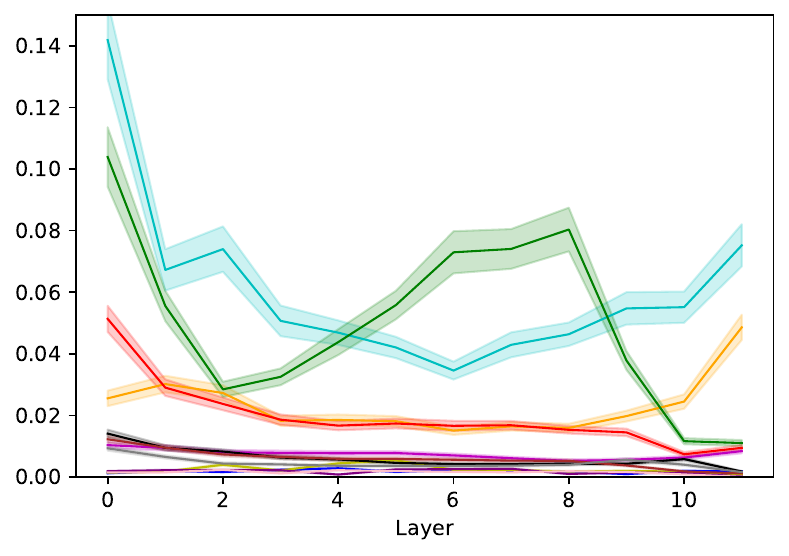}
    }
    \hfil
    \subfloat[$\|W_{fulllore (ours)} - W_{adversarial}\|$]{
      \includegraphics[width=0.41\textwidth]{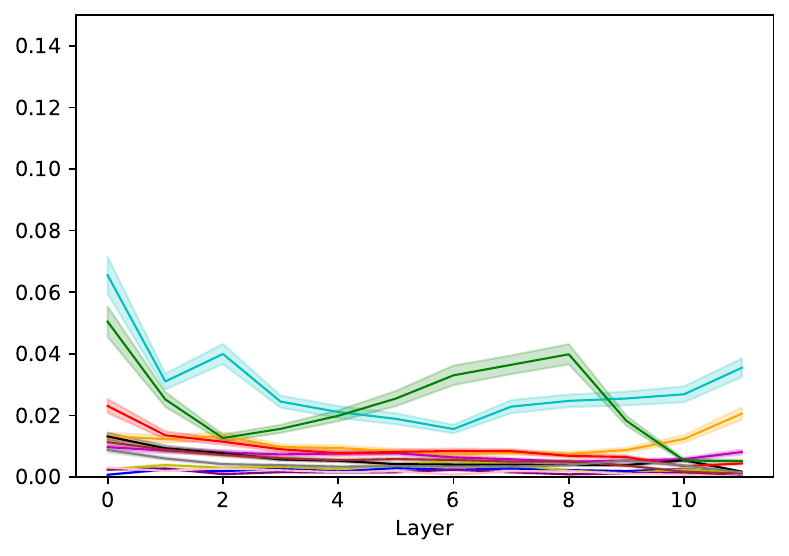}
    }
    \hfil
    \subfloat{
      \includegraphics[width=0.12\textwidth]{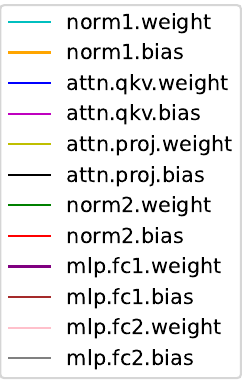}
    }
    \caption{Comparison of the distance between the parameters of different parameter-efficient training methods and the fully finetuned method across the 12 Transformer layers of the ViT model. (a) presents the distance between LoRA and fully finetuned model, and (b) presents the distance between our FullLoRA and fully finetuned model. The experiment is repeated by 5 times, the solid line represents the mean value of the 5 experiments, while the shaded area represents the standard deviation.}
    \label{fig:para_diff}
\end{figure*}

\subsection{Combined with Other Advanced Adversarial Training Methods}

Our method can also be further combined with other advanced adversarial training methods to enhance the results of our adversarial finetuning. The experiment is conducted using the ViT-B model on various datasets.

In~\cref{tab:combine_agat}, we present the results of combining FullLoRA with the AGAT method~\cite{wu2022towards}, an attention-guided tokens discarding approach designed to accelerate the speed of adversarial training.
Comparing our FullLoRA with the AGAT method, it is evident that our FullLoRA achieves significantly higher model robustness than the AGAT method, using only about 5\% of the training parameters during adversarial finetuning.
Furthermore, in~\cref{tab:combine_with_agat}, we also compare the results of the FullLoRA and LoRA methods when combined with the AGAT method, respectively. It is clear that our FullLoRA also achieves noticeably higher robustness than LoRA, fully demonstrating the superiority of our approach.

Additionally, we also combine our FullLoRA with other advanced methods aimed at enhancing the robustness of adversarial robustness. As shown in~\cref{tab:combine_performance}, when combined with MART~\cite{wang2020improving} and PRM~\cite{mo2022when}, our FullLoRA method still significantly outperforms the LoRA method in terms of model robustness across different datasets.

In summary, when combined with different advanced adversarial training methods, our FullLoRA consistently achieves higher model robustness on various datasets compared to the LoRA method, demonstrating the broad applicability of our proposed FullLoRA method.

\subsection{Comparison of the Distance Between Parameters of Different Methods and the Fully Finetuned Method}
\label{sec:distance}
Compared to LoRA method, our proposed FullLoRA method can more effectively narrow the parameter difference from the fully adversarial training model. To validate it, we conduct experiments to compare the parameter differences between models obtained through various parameter-efficient training methods (\eg, LoRA~\cite{hu2022lora} and our proposed FullLoRA) and models that have been fully finetuned. We denote the model obtained by finetuning a pre-trained model using LoRA as $W_{lora}$, that using our proposed FullLoRA as $W_{ours}$, and that from fully adversarial training as $W_{adversarial}$.
We integrate the parameters of the LoRA module into the weights of the Key, Query, and Value variables in the attention module, and the parameters of the layernorm layer in the LNLoRA module are merged into the layernorm layers preceding the MSA and MLP blocks.
We visualize  $\|W_{lora} - W_{adversarial}\| $ and $\|W_{ours} - W_{adversarial}\|$ of different parameters across the 12 Transformer layers of the ViT model. As shown in~\cref{fig:para_diff}, the visualizations reveal that the LoRA method does not effectively reduce the disparity between the pre-trained model and the adversarial model, whereas our proposed method more closely approximates the fully adversarially finetuned model, which well supports our claim that parameter disparities can not be adequately addressed by the existing LoRA method and our proposed LNLoRA aims to learn the differences in feature magnitude between standard and adversarial trained models.

\section{Conclusion}
\label{sec:conclusion}

In conclusion, to achieve parameter-efficient robustness finetuning for standard training models, our work presents the FullLoRA method, an innovative and lightweight approach designed to significantly enhance the adversarial robustness of pretrained ViT models in a parameter-efficient manner. By integrating the newly proposed LNLoRA modules into several key components of the ViT model, our method can rapidly improve the robustness of the pretrained model with a small number of extra parameters.  Extensive experiments show that our method not only addresses critical limitations of previous approaches but also showcases remarkable improvement in the robustness of models against adversarial attacks.

\bibliographystyle{IEEEtran}
\bibliography{main}

\newpage


\begin{IEEEbiography}[{\includegraphics[width=1in,height=1.25in,clip,keepaspectratio]{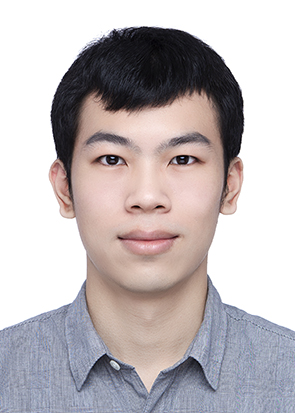}}]{Zheng Yuan}
    received the B.S. degree from University of Chinese Academy of Sciences in 2019. He is currently pursuing the Ph.D. degree from University of Chinese Academy of Sciences. His research interest includes adversarial example and model robustness. He has authored several academic papers in international conferences including ICCV/ECCV/ICPR and journals including TPAMI/IJCV.
  
\end{IEEEbiography}

\begin{IEEEbiography}[{\includegraphics[width=1in,height=1.25in,clip,keepaspectratio]{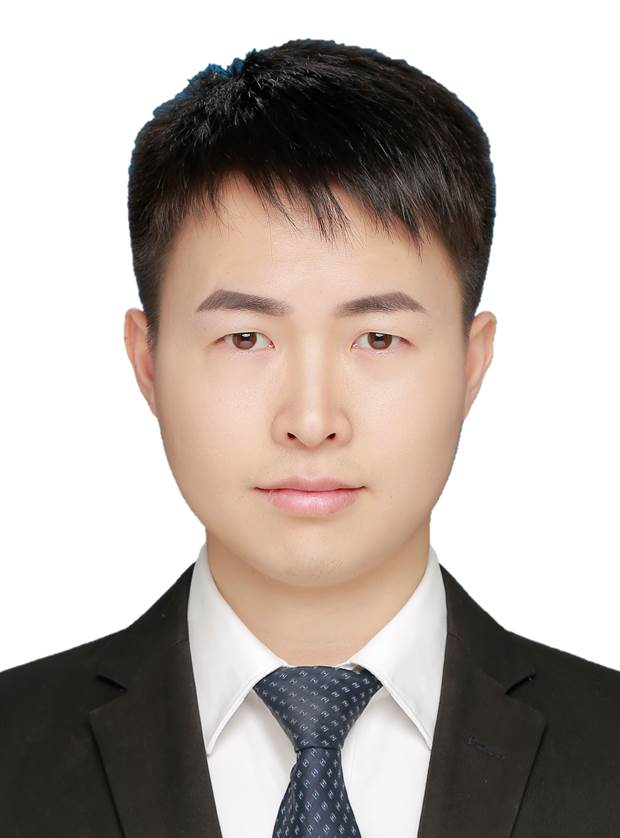}}]{Jie Zhang}
    (Member, IEEE) is an associate professor with the Institute of Computing Technology, Chinese Academy of Sciences (CAS). He received the Ph.D. degree from the University of Chinese Academy of Sciences, Beijing, China. His research interests cover computer vision, pattern recognition, machine learning, particularly include face recognition, image segmentation, weakly/semi-supervised learning, domain generalization.
\end{IEEEbiography}

\begin{IEEEbiography}[{\includegraphics[width=1in,height=1.25in,clip,keepaspectratio]{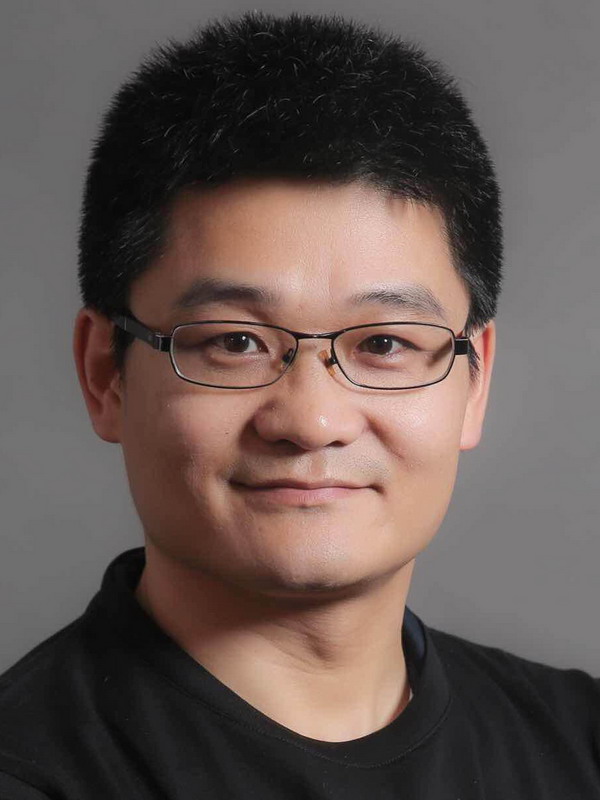}}]{Shiguang Shan}
    (Fellow, IEEE) received Ph.D. degree in computer science from the Institute of Computing Technology (ICT), Chinese Academy of Sciences (CAS), Beijing, China, in 2004. He has been a full Professor of this institute since 2010 and now the deputy director of CAS Key Lab of Intelligent Information Processing. His research interests cover computer vision, pattern recognition, and machine learning. He has published more than 300 papers, with totally more than 29,000 Google scholar citations. He has served as Area Chair (or Senior PC) for many international conferences including ICCV11, ICPR12/14/20, ACCV12/16/18, FG13/18, ICASSP14, BTAS18, AAAI20/21, IJCAI21, and CVPR19/20/21. And he was/is Associate Editors of several journals including IEEE T-IP, Neurocomputing, CVIU, and PRL. He was a recipient of the China's State Natural Science Award in 2015, and the China's State S\&T Progress Award in 2005 for his research work.
\end{IEEEbiography}
  
\begin{IEEEbiography}[{\includegraphics[width=1in,height=1.25in,clip,keepaspectratio]{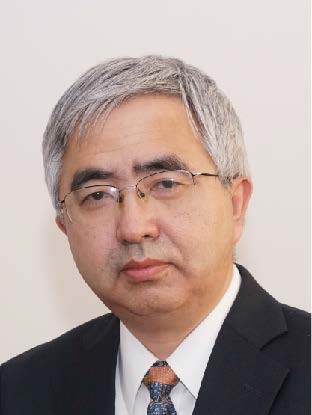}}]{Xilin Chen}
    (Fellow, IEEE) is currently a Professor with the Institute of Computing Technology, Chinese Academy of Sciences (CAS). He has authored one book and more than 400 articles in refereed journals and proceedings in the areas of computer vision, pattern recognition, image processing, and multimodal interfaces. He is a fellow of the ACM, IAPR, and CCF. He is also an Information Sciences Editorial Board Member of Fundamental Research, an Editorial Board Member of Research, a Senior Editor of the Journal of Visual Communication and Image Representation, and an Associate Editor-in-Chief of the Chinese Journal of Computers and Chinese Journal of Pattern Recognition and Artificial Intelligence. He served as an organizing committee member for multiple conferences, including the General Co-Chair of FG 2013/FG 2018, VCIP 2022, the Program Co-Chair of ICMI 2010/FG 2024, and an Area Chair of ICCV/CVPR/ECCV/NeurIPS for more than ten times. 
\end{IEEEbiography}

\vfill

\end{document}